\documentclass[letterpaper]{article} % DO NOT CHANGE THIS
\usepackage{aaai25}  % DO NOT CHANGE THIS
\usepackage{times}  % DO NOT CHANGE THIS
\usepackage{helvet}  % DO NOT CHANGE THIS
\usepackage{courier}  % DO NOT CHANGE THIS
\usepackage[hyphens]{url}  % DO NOT CHANGE THIS
\usepackage{graphicx} % DO NOT CHANGE THIS
\usepackage{natbib}  % DO NOT CHANGE THIS AND DO NOT ADD ANY OPTIONS TO IT
\usepackage{caption} % DO NOT CHANGE THIS AND DO NOT ADD ANY OPTIONS TO IT
\usepackage{algorithm}
\usepackage{algorithmic}

\usepackage{booktabs}
\usepackage{multirow}
\usepackage{amsfonts}
\usepackage{diagbox}
\usepackage{amsmath}
\usepackage{color}
\usepackage{subfigure}
\usepackage{amssymb}
\usepackage{mathtools}
\usepackage{amsthm}
\usepackage{bm}
\usepackage{physics}

\newcommand{\chen}[1]{\textcolor{blue}{\textbf{[minhan]:} #1}}

\usepackage{newfloat}

\usepackage{listings}
\DeclareCaptionStyle{ruled}{labelfont=normalfont,labelsep=colon,strut=off} % DO NOT CHANGE THIS
\floatstyle{ruled}
\newfloat{listing}{tb}{lst}{}
\floatname{listing}{Listing}
\title{Hierarchical Gradient-Based Genetic Sampling for Accurate Prediction of Biological Oscillations}
\author{
    Heng Rao\textsuperscript{\rm 1},
    Yu Gu\textsuperscript{\rm 1},
    Jason Zipeng Zhang\textsuperscript{\rm 2},
    Ge Yu\textsuperscript{\rm 1},
    Yang Cao\textsuperscript{\rm 3},
    Minghan Chen\textsuperscript{\rm 2}*
}
\affiliations{
    \textsuperscript{\rm 1}College of Computer Science and Engineering, Northeastern University, Shenyang, China\\
    \textsuperscript{\rm 2}Department of Computer Science, Wake Forest University, Winston-Salem, NC, USA\\
    \textsuperscript{\rm 3}Department of Computer Science, Virginia Polytechnic Institute and State University, Blacksburg, VA, USA\\
    *Corresponding author: chenm@wfu.edu
}

\usepackage{bibentry}
\begin{document}

\maketitle

%%% Abstract
\begin{abstract}
Biological oscillations are periodic changes in various signaling processes crucial for the proper functioning of living organisms. These oscillations are modeled by ordinary differential equations, with coefficient variations leading to diverse periodic behaviors, typically measured by oscillatory frequencies. This paper explores sampling techniques for neural networks to model the relationship between system coefficients and oscillatory frequency. However, the scarcity of oscillations in the vast coefficient space results in many samples exhibiting non-periodic behaviors, and small coefficient changes near oscillation boundaries can significantly alter oscillatory properties. This leads to non-oscillatory bias and boundary sensitivity, making accurate predictions difficult. While existing importance and uncertainty sampling approaches partially mitigate these challenges, they either fail to resolve the sensitivity problem or result in redundant sampling. To address these limitations, we propose the Hierarchical Gradient-based Genetic Sampling (HGGS) framework, which improves the accuracy of neural network predictions for biological oscillations. The first layer, Gradient-based Filtering, extracts sensitive oscillation boundaries and removes redundant non-oscillatory samples, creating a balanced coarse dataset. The second layer, Multigrid Genetic Sampling, utilizes residual information to refine these boundaries and explore new high-residual regions, increasing data diversity for model training. Experimental results demonstrate that HGGS outperforms seven comparative sampling methods across four biological systems, highlighting its effectiveness in enhancing sampling and prediction accuracy.

\end{abstract}

% Uncomment the following to link to your code, datasets, an extended version or similar.
%
% \begin{links}
%     \link{Code}{https://aaai.org/example/code}
%     \link{Datasets}{https://aaai.org/example/datasets}
%     \link{Extended version}{https://aaai.org/example/extended-version}
% \end{links}
\section{1 Introduction}\label{sec:intro}
% \cite{guo2008class}
%%% first
In biological systems, oscillation refers to the repetitive, cyclical behaviors of cell signaling and biological processes over time, such as fluctuations in concentrations of biochemical substances \cite{tamate2017effect}, rhythmic activities in cellular functions \cite{goldbeter2002computational}, or periodic changes in physiological states \cite{kurosawa2002comparative}. Understanding these oscillations is crucial for comprehending how biological systems maintain stability, respond to external stimuli, and regulate complex processes. 
To study oscillations, researchers often use ordinary differential equations (ODEs) to model the temporal dynamics that characterize oscillatory patterns in biological systems.
Recently, the application of machine learning to biological systems has gained significant attention. In many studies \cite{daneker2023systems,yazdani2020systems,szep2021parameter}, system-level mathematical models of biological reactions are integrated with machine learning models to establish relationships between raw data and system coefficients.
Of these, Neural Networks (NNs) show promising results, offering an effective tool for predicting biological oscillations based on system coefficients. 

\begin{figure}[!tb]
    \centering
    \includegraphics[width=\linewidth]{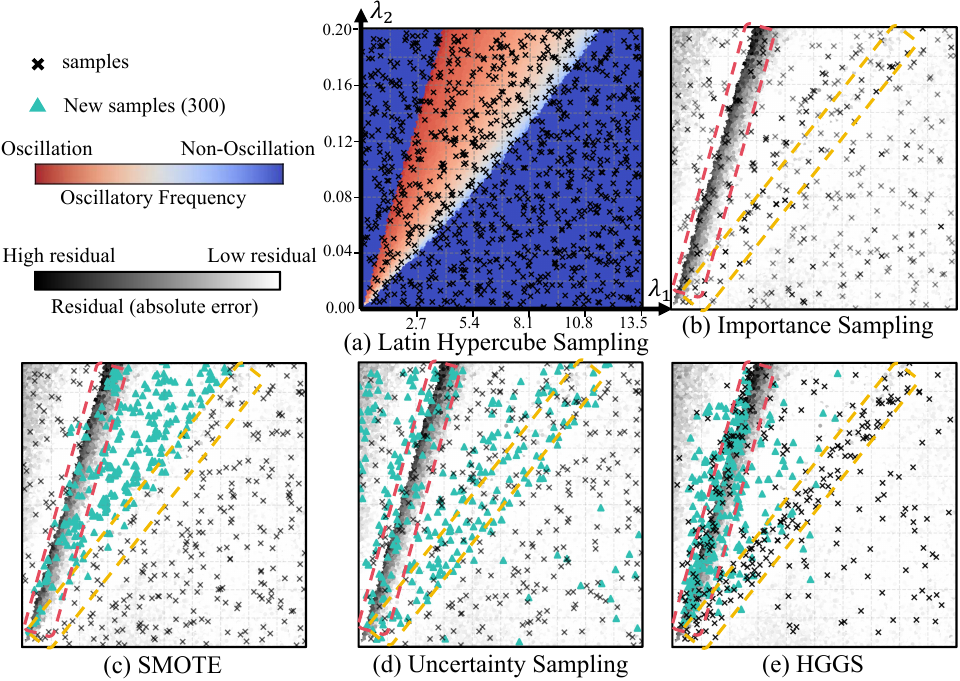}
    \caption{Examples of different sampling methods applied to the system coefficients of a cell cycle model \cite{liu2012hybrid}. The orange-blue and black-white color bars indicate trends in oscillation and residual changes, respectively. (a) LHS performs a random sampling of 1000 points in the coefficient domain. (b) IS fails to introduce new samples; (c-d) SMOTE and US lack sufficient coverage in high residual regions (red box). US also produces redundant samples spreading across the domain; (e) Our proposed HGGS method extracts the boundary information (red, yellow boxes) and effectively generates new samples (tear triangle) concentrating on high-residual regions (red box).}
    \label{fig:samplingmethod}
\end{figure}

However, the relationship between system coefficients and oscillatory behaviors is highly complex \cite{stark2007oscillations}. Variations in these coefficients can disrupt oscillations, causing non-oscillatory states. 
In biological systems, coefficients that can lead to oscillatory states are relatively rare; most coefficient combinations result in non-oscillatory behavior, especially in high-dimensional domains. The prevalence of non-oscillatory states creates a significant data imbalance, challenging the prediction of oscillatory frequency---a key characteristic of oscillation, defined as the inverse of the oscillatory period ($\frac{1}{period}$). When oscillation is absent, the frequency is set to zero.
Fig.~\ref{fig:samplingmethod}(a) illustrates the mapping between system coefficients and oscillatory frequencies of a cell cycle system \cite{liu2012hybrid}. The domain shows data imbalance with 70\% non-oscillation (blue).
Moreover, minor changes in these coefficients can lead to substantial changes in oscillatory frequencies, particularly in boundary areas where oscillatory states transition to non-oscillatory states. This heightened sensitivity further complicates accurate predictions and is reflected in the poor performance of NNs in these regions, as evidenced by high residuals (absolute error) marked as black shades in Fig.~\ref{fig:samplingmethod}. 
Overall, we face two challenges: 
\begin{itemize}
    \item \textbf{Non-Oscillatory Bias.} The majority of data samples exhibit non-oscillatory behaviors, resulting in an abundance of redundant information that does not improve model accuracy and may even reduce efficiency. %\cite{toneva2018empirical}.
    \item \textbf{Oscillatory Boundary Sensitivity.} Sharp transitions between non-oscillatory and oscillatory states occur in narrow boundaries of the coefficient space, which are easily overlooked by random sampling and lead to high errors.
\end{itemize}
% However, as shown in Fig.~\ref{fig:samplingmethod}, varying parameters $\lambda_1$ and $\lambda_2$ predominantly leads to non-oscillatory behavior (dark blue region), which is irrelevant for biological studies. Based on the consideration of the data domain for this problem, we found that there is a data imbalance issue in the biological system dataset. It will result in poor performance of the model in predicting oscillatory frequency, which is our primary concern \cite{longadge2013multi}. Additionally, we found that small variations in the $\lambda_1$ cause drastic changes in the frequency (namely boundary), leading to data sensitivity problem in traditional random sampling method \cite{gong2019diversity}. Experiments demonstrated that the machine learning model performs worst in these sensitive regions. In summary, using random sampling to obtain the dataset for training model will lead to the following two issues:
%%%% the method of solve these problem(related work)
Therefore, a more effective sampling strategy is necessary to obtain more balanced and representative samples for model training. 
%\cite{marchant2017search,lu2023pa,panigrahi2024comparing} 
Importance Sampling (IS) \cite{liu2021low,lu2023pa} can be used to resample more minority class instances, balancing the majority-minority ratio. As shown in Fig.~\ref{fig:samplingmethod}(b), IS resamples the oscillation class, with darker samples indicating higher resampling, but it cannot generate new samples to alleviate boundary sensitivity issues. 
Synthetic Minority Oversampling Technique (SMOTE) \cite{chawla2002smote,torgo2013smote} offers another way to balancing the dataset by generating synthetic minority samples. However, it insufficiently covers high-residual areas (red box), see Fig.~\ref{fig:samplingmethod}(c). Targeting data sensitivity, Uncertainty Sampling (US) \cite{liu2023understanding} selects informative samples based on predicted importance. However, due to limited diversity in the random selection process, US often produces redundant samples, reducing efficiency. This is evident in Fig.~\ref{fig:samplingmethod}(d), where new samples are scattered across the domain but fail to adequately cover high-residual regions (red box) compared to our method in Fig.~\ref{fig:samplingmethod}(e). Thus, these sampling strategies fail to effectively address the challenges.

%%%% our work(Brief)
This work introduces the Hierarchical Gradient-based Genetic Sampling (HGGS) method, a novel two-layer framework designed to enhance the effectiveness of selecting representative samples and improve the accuracy of predictions for biological oscillations.
%%This approach enables the model to effectively capture the complex relationship between the system coefficients and biological oscillations. 
Specifically, starting with an initial candidate set obtained by Latin Hypercube Sampling (LHS), we first apply Gradient-based Filtering (GF) to select samples near boundary regions and balance the proportions of non-oscillation (majority) and oscillation (minority) samples. This GF layer yields a balanced coarse dataset that guides subsequent refinement. In the second layer, Multigrid Genetic Sampling (MGS) dynamically constructs grids at multiple levels based on residual (absolute error) information from training data. By sampling within these grids, MGS not only enriches existing boundaries but also explores new high-residual areas. Through continual learning, HGGS iteratively refines the training dataset with more representative samples, consistently improving model performance. The main contributions of this paper are summarized below:
% \begin{itemize}
%     \item \textbf{Identification of Boundary Areas:} The simple and efficient Gradient-based Filtering method identifies boundary samples in the dataset and addresses the issue of data imbalance between majority (non-oscillatory) and minority (oscillatory) classes.
%     \item \textbf{Generation of New Samples:} The proposed multi-grid Genetic Sampling method guided by the residual distribution of the training dataset. generates new samples, especially in unexplored boundary areas, thereby increasing the diversity of the dataset.
%     \item \textbf{Improved Performance:} Our experiments show that the proposed method achieves lower error compared to state-of-the-art methods and ensures diverse, balanced training data across four biological systems.
% \end{itemize}

% Modified
\begin{itemize}
    \item We propose a simple and efficient Gradient-based Filtering technique that can extract oscillation boundaries, which often entail high residuals, and removes redundant non-oscillatory samples. The GF layer generates balanced coarse data, enabling efficient MGS refinement. 
    \item Our Multigrid Genetic Sampling strategy leverages residual information to refine existing boundaries and explore new high-residual regions. The MGS layer systematically samples across grids at different levels, reducing sensitivity and enhancing oscillation diversity.
    \item The proposed HGGS method ensures a representative dataset through continuous adaption to the evolving error landscape during training, showing superior accuracy over seven baselines across four biological systems. HGGS is versatile and can be applied to predict various systematic features beyond biological oscillations.
\end{itemize}

%dynamically adapts to the evolving error landscape, ensuring that the most challenging regions are accurately represented in the training data. 

% Gradient-based Filtering combined with Genetic Sampling can address data imbalance and lack of data diversity problems. Gradient-based Filtering retains informative, potentially hard-to-learn samples and balances the proportion of Non-Oscillation (majority class) and Oscillation (minority class) data. Additionally, Gradient-based Filtering serves as a fundamental step for Genetic Sampling, which can efficiently and quickly sample new informative data. 
%%% The following paper structure

\section{2 Related Work}\label{sec:relatedwork}

% \subsection{Static Sampling}
\textbf{Sampling for Data Imbalance.}
Data imbalance often results in biased model performance, favoring majority samples while underperforming on minority ones. To address this issue, several sampling techniques have been proposed, including undersampling, oversampling, and importance sampling techniques. 
% Modified [1]
% Undersampling randomly removes majority samples \cite{guo2008class} to balance the majority-minority ratio, but it risks discarding informative data. 
% Modified [2]
Undersampling randomly removes majority samples \cite{wilson1972asymptotic,tomek1976two,guo2008class} to balance the imbalance ratio, but it risks discarding informative data. 
On the other hand, oversampling mitigates data imbalance by augmenting minority class samples, as in SMOTE \cite{chawla2002smote, torgo2013smote}. While SMOTE generates new minority samples to improve balance, it can introduce noise due to interpolated labels and may struggle to create truly representative minority samples. Importance Sampling (IS) \cite{liu2021low,lu2023pa} combines elements of both undersampling and oversampling by resampling more informative minority samples and excluding well-performing majority samples. However, it is prone to overfitting due to resampling and does not incorporate new data. Although these techniques offer various ways to alleviate data imbalance, they all struggle with the data sensitivity challenge.

\subsubsection{Dynamic Sampling.}
Dynamic sampling offers a more adaptive approach to selecting informative samples, enhancing model training by adjusting sample selection based on the model's evolving state.
This idea is widely utilized in Active Learning (AL), where various dynamic sampling methods leverage direct or indirect information from the model to select the most representative samples from unlabeled data.
One prominent method in AL is Uncertainty Sampling (US) \cite{lewis1994heterogeneous,5272205,liu2023understanding}, which selects samples based on their uncertainty scores. In classification, these scores can be calculated using techniques such as entropy uncertainty \cite{shannon1948mathematical} or confidence margin uncertainty \cite{sharma2017evidence}. 
Diversity-based strategies aim to select a broad range of samples based on data distribution, employing methods like gradient representation \cite{saran2023streaming} and switch events \cite{benkert2023gaussian}. Query by committee methods \cite{burbidge2007active,kee2018query,hino2023active} aggregate outputs from multiple models to form new discriminative criteria, identifying the most representative samples for labeling by considering the underlying data distribution.
However, uncertainty sampling and diversity-based sampling often introduce significant redundancy, reducing sampling efficiency. Additionally, most of these methods are designed for nominal target variables and are rarely applicable to regression problems with continuous targets \cite{liu2023understanding}. 
% In contrast, our approach employs genetic sampling to consistently select the most representative samples from the labeled set for subsequent training, effectively mitigating data imbalance and sensitivity issues while maintaining sampling efficiency.
% Modified
In contrast, our HGGS method leverages residual information to ensure targeted sampling in high-residual areas and avoid redundancy, boosting both effectiveness and efficiency. 

\section{3 Preliminary}\label{sec:preliminary}
In this paper, we utilize a neural network model $\hat{y}(\bm{\lambda})=f_{nn}(\bm{\lambda};\bm{\Theta})$ to approximate the oscillatory frequency $y(\bm{\lambda})=f_P(\bm{u}(\bm{t},\bm{\lambda}))\in \mathbb{R}^{D'}$ of a system under initial state $\bm{u}_0$, given any set of coefficients $\bm{\lambda}\in \mathbb{R}^D$ in the ODEs presented below.
\begin{equation}\label{ODEs}
\begin{aligned}
\dv{\bm{u}}{t}&=\bm{\mathcal{N}}(\bm{u}),t\in[0,T] \\
\bm{u}(t,\bm{\lambda})|_{t=0}&=\bm{u}_0(\bm{\lambda}),\bm{\lambda}\in\Omega,
\end{aligned}
\end{equation}
where $f_P$ is the oscillatory operator used to calculate the oscillatory frequency $f_P(\bm{u}(\bm{t}, \bm{\lambda}))$ of $\bm{u}$ over the time span $\bm{t} = [0, t_1, t_2, \ldots, T]^\top$ using \cite{APICELLA201330}. $\bm{\mathcal{N}}(\bm{u})$ denotes a nonlinear operator consisting of variables vector $\bm{u}$, $\bm{\lambda} \in \Omega$ is a $D$-dimensional system coefficient vector, $\Omega$ is a subset of $\mathbb{R}^D$, $\bm{u}_0$ is the initial condition vector in $D'$ dimensions, and $\bm{\Theta}$ is the parameter of the neural network. Our goal is to minimize the error in approximating the original function $f_P(\bm{u}(\bm{t},\bm{\lambda}))$ using a neural network $f_{nn} \in F : \mathbb{R}^D \rightarrow \mathbb{R}^{D'}$, where $F$ represents an appropriate function mapping space. 
% Modified [1]
We achieve this by employing supervised learning with the following loss function.
\begin{equation}\label{loss function(continue)}
L(f_{nn}(\bm{\lambda}; \bm{\Theta}),y)=\norm{l(f_{nn}(\bm{\lambda}; \bm{\Theta}),y)}_{2,\Omega}^2, 
\end{equation}
where $l(f_{nn}(\bm{\lambda}; \bm{\Theta}),y) = \abs{f_{nn}(\bm{\lambda}; \bm{\Theta}) - y}$ and $\norm{\cdot}_2$ denotes Euclidean normalization.

In practice, Monte Carlo (MC) approximation is used to estimate the overall error $L$ using a finite set of sampled points from the domain, denoted as $L_N$. Thus, the loss function can be approximated as:
\begin{equation}\label{loss function}
\begin{aligned}
L(\bm{\Theta}) \approx L_N(\bm{\Theta})
&=\norm{l(f_{nn}(\bm{\lambda}; \bm{\Theta}),y)}_{2,S_\Omega}^2 \\
&=\frac {1} {N}\sum_{i=1}^{N}\norm{l(f_{nn}(\bm{\lambda}^{(i)}; \bm{\Theta}),y^{(i)})}_2^2,
\end{aligned}
\end{equation}
where $N$ is the number of samples and $S_\Omega = \{\bm{\lambda}^{(i)}\}_{i=1}^{N}$ represents the set of sampled coefficients from $\Omega$. To theoretically prove that sampling methods can further reduce the error between $f_{nn}$ and $f_P$, we build on the framework from \cite{tang2023adversarial}. 
Through analyzing the MC approximate loss function $L_N$, we derive the theoretical upper bound of the optimal model $f_{nn}(\cdot)_N^*$ approximating to oscillatory frequency $y(\cdot)$. Additionally, we introduce the theoretically optimal model $f_{nn}(\cdot)^*$ from overall error $L$, categorizing the error into two parts:
\begin{equation*}\label{upper bound loss function}
\begin{aligned}
    E(||f_{nn}(\cdot)_N^*-y(\cdot)||_\Omega)
    \leq &E(||f_{nn}(\cdot)_N^*-f_{nn}(\cdot)^*||_\Omega)\\
    &+||f_{nn}(\cdot)^*-y(\cdot))||_\Omega .
\end{aligned}
\end{equation*}
Here, $E$ is the expectation operator, and $\norm{\cdot}_\Omega$ is the norm operator in the function space $F$. 
The first term represents the statistical error, while the second term reflects the model approximation error determined by model structure. Sampling new data from the residual distribution over $\Omega$ can reduce statistical error. Proof details are provided in the Appendix.

\section{4 Method}\label{sec:method}

%%%%%%% Figure2 Architecture
\begin{figure*}[!tb]
% \begin{center}
\includegraphics[width=\linewidth]{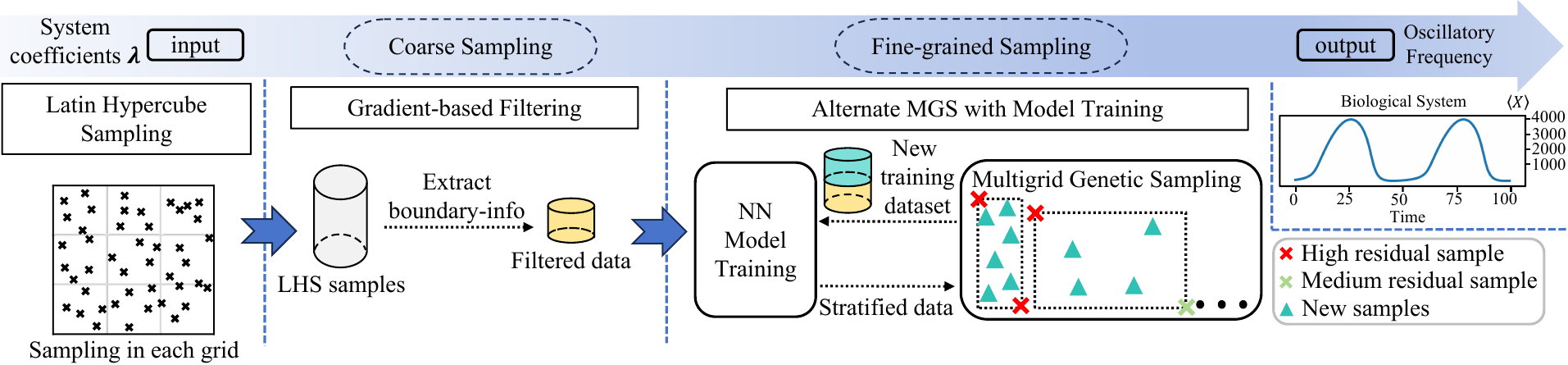}
% \caption{(a) An oscillatory sample($\lambda_1$=1.35; $\lambda_2$=0.02), characterized by a frequency label of 0.00826 for the variable ⟨~$X$~⟩. The plot illustrates the temporal dynamics of variable ⟨~$X$~⟩ over 1000 time steps under default cell cycle system coefficient, demonstrating a periodic pattern. (b) A non-oscillatory sample ($\lambda_1$=11.76; $\lambda_2$=0.16), with a manually set frequency label of 0. The variable ⟨~$X$~⟩ remains essentially constant over the course of 1000 time steps.}
\caption{Overview of Hierarchical Gradient-based Genetic Sampling framework: The input consists of initial samples obtained via LHS. Gradient-based Filtering is then employed to extract sensitive boundary information and eliminate redundant samples, generating balanced coarse data. Next, Multigrid Genetic Sampling constructs candidate sampling from stratified data to sharpen boundary precision. During training, new instances in high-residual areas are explored to continuously improve model performance. Finally, given any set of coefficients $\bm{\lambda}$, the model predicts the oscillatory frequency of the biological system.}
\label{fig:Architecture}
% \end{center}
\end{figure*}

This section describes the Hierarchical Gradient-based Genetic Sampling (HGGS) framework (Fig.~\ref{fig:Architecture}), which divides the sampling process into two layers: Gradient-based Filtering and Multigrid Genetic Sampling. By employing HGGS, we obtain a diverse and balanced dataset, significantly reducing model statistical error.

\subsection{Gradient-based Filtering}
% \ref{Architecture}
% Inspired by \cite{toneva2018empirical}, 
As demonstrated in Fig.~\ref{fig:samplingmethod}, biological systems often contain a significant amount of redundant data in non-oscillatory regions. Removing these redundant non-oscillatory samples not only preserves the model's overall performance but also increases the proportion of oscillatory samples in the dataset, thereby mitigating the effects of non-oscillatory bias.

To achieve this, we design an effective Gradient-based Filtering (GF) technique that identifies sensitive regions, particularly boundary areas where small changes can lead to significant shifts in oscillatory frequency (see red, yellow boxes in Fig.~\ref{fig:samplingmethod}(e)), which are also characterized by higher residuals. In this context, we first rank all samples from the initial dataset $S_\Omega$ using Eq. \ref{samples_gradient}. 
\begin{equation}\label{samples_gradient}
\begin{aligned}
gd(\bm{\lambda}^{(i)})_{S_\Omega}=\frac {1} {K} \sum_{j=1}^{K}\frac {\norm{y^{(j)}-y^{(i)}}_2^2} {\norm{\bm{\lambda}^{(j)}-\bm{\lambda}^{(i)}}_2^2};\ 
\bm{\lambda}^{i}, \bm{\lambda}^{j}\in S_{\Omega}, i\neq j,
\end{aligned}
\end{equation}
where $gd(\bm{\lambda}^{(i)})_{S_\Omega}$ represents the gradient degree of collocation sample $i$ from the training dataset $S_\Omega$ and $K$ is a hyperparameter denoting the top-$K$ nearest collocation samples to $i$ in the training dataset $S_\Omega$. This function assesses the importance of collocation sample $i$; a higher $gd(\bm{\lambda}^{(i)})_{S_\Omega}$ value indicates that the sample is closer to boundary areas.

According to this gradient ranking, we filter the top-$r\%$ samples $S_{\Omega_{gf1}}$ from $S_\Omega$, where $r$ is the filtering ratio to extract the boundary information. 
Next, to enhance the model's adaptability to non-oscillatory conditions, we retain an essential global dataset, $S_{\Omega_{gf2}}$, from the remaining samples in $S_{\Omega} - S_{\Omega_{gf1}}$ using uncertainty sampling \cite{liu2023understanding}. This subset is specifically selected to capture the overall characteristics and maintain performance for non-oscillatory regions. We then formulate the coarse training set $S_{\Omega}^{(0)} = S_{\Omega_{gf}} = S_{\Omega_{gf1}} \cup S_{\Omega_{gf2}}$, where $S_{{\Omega}_{gf}} \subseteq S_\Omega$ and $|S_{{\Omega}_{gf}}| < |S_\Omega|$. In $S_{{\Omega}_{gf}}$, most redundant non-oscillation samples from the original dataset are removed, while a diverse set of oscillatory data is retained. 
\subsection{Multigrid Genetic Sampling}
From the prior GF layer, we obtained a balanced coarse set of informative samples $S_\Omega^{(0)}$. Our next goal is to sharpen boundary precision and explore new high-residual areas through effective sampling. To achieve this, constructing an accurate residual distribution is essential. Since non-oscillatory regions correspond to low probabilities in the distribution, and therefore contribute minimally to model performance. Our focus is on constructing high-residual distribution. 

However, the small sample set $S_\Omega$ obtained through random sampling does not sufficiently represent the high-residual distribution $p_g$ over the domain $\Omega$. 
To address this limitation, we construct multiple sampling grids of varying sizes to capture the high-residual distribution $p_g$.

We first use a Gaussian Mixture model \cite{reynolds2009gaussian} to characterize three potential residual distributions within $S_\Omega$: low-, medium-, and high-residual distributions, respectively, denoted as $S_{{\Omega}_{lr}}$, $S_{{\Omega}_{mr}}$, and $S_{{\Omega}_{hr}}$. 
% Specifically, the construction of these grids is based on the assumption that high-residual regions exhibit locality. 
% By excluding low-residual sample set $S_{{\Omega}_{lr}}$ from the initial set $S_\Omega$, we increase the chances of high-residual samples being included in grid construction, thus better representing the high-residual distribution over $\Omega$.
Next, each grid is defined as a hypercube bounded by points $\bm{\lambda}^{(i)}$ and $\bm{\lambda}^{(j)}$, both originating from $S_\Omega^{(0)}-S_{{\Omega}_{lr}}^{(0)}$.
To ensure comprehensive coverage of the high-residual distribution, it is necessary to construct a large number of grids at different sizes, each composed of diverse samples.

Building on the idea of multigrid, we integrate genetic sampling into a method called Multigrid Genetic Sampling (MGS) to approximate the high-residual distribution over $\Omega$ and further refine these areas. Specifically, we sample new points within each hypercubic grid according to Eq. \ref{Sampling in Grid}:
\begin{equation}\label{Sampling in Grid}
\begin{aligned}
    \bm{\lambda}_{new}=\alpha \odot (\bm{\lambda}^{(i)}-\bm{\lambda}^{(j)})+\bm{\lambda}^{(j)};\ 
    \alpha\in [0,1]^D,i\neq j,
\end{aligned}
\end{equation}
where $\alpha$ is a D-dimensional weight vector with each value in the range [0,1], and $\odot$ denotes component-wise multiplication. 
$\bm{\lambda}_{new}$ is a randomly sampled point within the hypercubic enclosed by $\bm{\lambda}^{(i)}$ and $\bm{\lambda}^{(j)}$.

% However, constructing a multigrid from a small sample $S_\Omega^{(0)}$ may not accurately capture all possible high-residual areas of $\Omega$. 
In order to obtain a fine-grained dataset from $S_\Omega^{(0)}$, we alternate GMS with model training to continuously refine the high-residual areas. We define $m_c$ as the number of sampling cycles and $m_e$ as the number of training epochs. The $k'$-th sampled dataset by GMS is denoted as $S^{(k')}_{{\Omega}_{gs}}$, and the $k$-th training dataset is $S^{(k)}_\Omega = \bigcup_{k'=1}^{k} S^{(k')}_{{\Omega}_{gs}} \cup S_{{\Omega}_{gf}}$.

For the ($k+1$)-th sampling, we utilize the current $k$-th residual samples, categorized in low-, medium-, and high-residual distributions, denoted as  $S^{(k)}_{{\Omega}_{lr}}$, $S^{(k)}_{{\Omega}_{mr}}$, and $S^{(k)}_{{\Omega}_{hr}}$.
% we use a Gaussian Mixture \cite{reynolds2009gaussian} model to characterize three potential residual distributions within $S_\Omega^{(k)}$: low, medium, and high-residual distributions, respectively denoted as $S^{(k)}_{{\Omega}_{lr}}$, $S^{(k)}_{{\Omega}_{mr}}$, and $S^{(k)}_{{\Omega}_{hr}}$. 
Next, to exploit and explore the high-residual domain, we defined crossover and mutation operations over medium- and high-residual sets as follows:
\begin{itemize}
    \item \text{Multigrid Crossover}: Randomly select two distinct samples $\bm{\lambda}^{(i)}_{hr}$ and $\bm{\lambda}^{(j)}_{hr}$ from the high-residual subset $S^{(k)}_{{\Omega}_{hr}}$, and
    % Within the hypercubic grid formed by the two samples, 
    randomly sample a new point $\bm{\lambda}_{hh}$ by Eq. \ref{Sampling in Grid} and add it to the sample set $S^{(k+1)}_{{\Omega}_{gs}}$.
    This operation serves to exploit and refine existing high-residual areas.

    \item \text{Multigrid Mutation}: Randomly select $\bm{\lambda}^{(i)}_{hr}$ from $S^{(k)}_{{\Omega}_{hr}}$ and $\bm{\lambda}^{(j)}_{mr}$ from $S^{(k)}_{{\Omega}_{mr}}$. Form a hypercubic grid using the two samples and sample a new point $\bm{\lambda}_{hm}$ by Eq. \ref{Sampling in Grid}. Add this new sample to $S^{(k+1)}_{{\Omega}_{gs}}$.
    The mutation operation serves to explore global boundaries and new high-residual areas.
\end{itemize}
Before the $(k+1)$-th training round, our method samples $n_{v1}$ and $n_{v2}$ points using the aforementioned operations, such that $|S^{(k+1)}_{{\Omega}_{gs}}| = n_{v1} + n_{v2}$ and $S^{(k+1)}_{{\Omega}_{gs}} = \bigcup_{i=1}^{n_{v1}} \bm{\lambda}^{(i)}_{hh} \cup \bigcup_{i=1}^{n_{v2}} \bm{\lambda}^{(i)}_{hm}$. Thus, the sample set for the $(k+1)$-th training round is $S^{(k+1)}_\Omega = S^{(k+1)}_{{\Omega}_{gs}} \cup S^{(k)}_\Omega$. At the end of sampling, $n_s=m_c\times(n_{v1}+n_{v2})$ and $S_{\Omega_{gs}}=\bigcup_{k=1}^{m_c} S^{(k)}_{\Omega}$.

The GF layer followed by the GMS layer, allows for effective and efficient sampling within high residuals. The pseudocode is presented in Algorithm \ref{alg:HGGS}.

\begin{algorithm}[tb]
\caption{HGGS for predicting biological oscillations}
\label{alg:HGGS}
\textbf{Input}: NN model $f_{nn}(\bm{\lambda}; \bm{\Theta})$, neighbors for gradient estimation $K$, initial sample size $N$, filtering ratio $r$, sampling cycle $m_c$, Multigrid Genetic Sampling budget $\{n_{v1}, n_{v2}\}$\\
\textbf{Initialization}: LHS $S_{\Omega}=\{(\bm{\lambda}^{(i)}, y^{(i)})\}_{i=1}^{N}$\\
\textbf{Output}: Target model $f_{nn}(\bm{\lambda}; \bm{\Theta}^*)$
\begin{algorithmic}[1] %[1] enables line numbers
\STATE Apply Gradient-based Filtering to $S_{\Omega}$ to generate $S_{\Omega}^{(0)}$: $S_{\Omega}^{(0)} \leftarrow GF(S_{\Omega}, r, K)$ where $S_{\Omega}^{(0)} \subseteq S_{\Omega}$
\STATE Update $f_{nn}(\bm{\lambda}; \bm{\Theta})$ by minimizing $L_{|S_{\Omega}^{(0)}|}$ in Eq. \ref{loss function}
\FOR{$k = 0, 1, \ldots, m_c - 1$}
\STATE Compute residual $l = \{|f_{nn}(\bm{\lambda}^{(i)}; \bm{\Theta}^{(k)})-y^{(i)}|\}_{i=1}^{|S_{\Omega}^{(k)}|}$
\STATE Stratify $S_{\Omega}^{(k)}$ into 3 subdomains based on residual using Gaussian Mixture: $\{S^{(k)}_{\Omega_{lr}}, S^{(k)}_{\Omega_{mr}}, S^{(k)}_{\Omega_{hr}}\}$
\STATE $S_{\Omega_{gs}}^{(k+1)} \leftarrow MGS(n_{v1}, n_{v2}, S^{(k)}_{\Omega_{mr}}, S^{(k)}_{\Omega_{hr}})$
\STATE $S_{\Omega}^{(k+1)} \leftarrow S_{\Omega}^{(k)} \cup S_{\Omega_{gs}}^{(k+1)}$    // Update datasets
\STATE Update $f_{nn}(\bm{\lambda}; \bm{\Theta})$ by minimizing $L_{|S_\Omega^{(k+1)}|}$ in Eq. \ref{loss function}
\ENDFOR
\STATE \textbf{return} $f_{nn}(\bm{\lambda}; \bm{\Theta}^*)$
\end{algorithmic}
\end{algorithm}
\section{5 Experiments}\label{sec:evaluation}
To validate the proposed HGGS, we conducted experiments on four biological systems known for their oscillatory behaviors. We also performed an ablation study to assess Gradient-based Filtering and Multigrid Genetic Sampling, and examined the sensitivity of model hyperparameters.

\subsection{Biological System Dataset}
Benchmark datasets from four biological systems were used for method evaluation: the Brusselator system \cite{prigogine1978time}, the Cell Cycle system \cite{liu2012hybrid}, the Mitotic Promoting Factor (MPF) system \cite{novak1993modeling}, and the Activator Inhibitor system \cite{murray2002mathematical}. 
For each system, we generated 20k--70k sets of system coefficients using LHS, ran simulations to produce system dynamics, and determined the oscillatory frequency using \cite{APICELLA201330}.
This data was then used for training and testing of our method.
Descriptions of the four biological systems, their corresponding ODEs, and detailed simulation settings are provided in the Appendix.

\subsection{Baselines}
The proposed HGGS was compared with seven baselines:
%Except for LHS, the training process alternates with data refinement within the continual learning framework.
\begin{itemize}
    \item Latin Hypercube Sampling (LHS) \cite{stein1987large}: Divides the coefficient domain into equal grids and randomly samples from each grid to ensure full coverage.
    \item Importance Sampling (IS) \cite{lu2023pa}: Resamples based on residual information, updating the training set after each epoch. IS$^\dag$ is an optimization that updates only when the model shows no improvement.
    % The sampling distribution gradually shifts from the initial random distribution to the residual information distribution of the samples. This is achieved through resampling to update the training samples. We implemented two versions: the vanilla version (resample each epoch) denoted as IS, and the optimized version (resample when the last set of samples has converged) denoted as IS$^\dag$.
    % \item Synthetic Minority Over-sampling Technique (SMOTE) \cite{torgo2013smote}: To alleviate data imbalance, new samples are constructed from existing sample labels and added to the initial training set. Here, the label information is obtained through querying rather than pseudo-labeling.
    \item Uncertainty Sampling (US) \cite{liu2023understanding}: Selects the top-$o$ new samples from a candidate set based on residual distribution, and adds them to the training set. We implemented the candidate set in two ways: pool-based (US-P) and streaming-based (US-S).
    \item Weight Reservoir Sampling (WRS) \cite{efraimidis2006weighted}: Selects the top-$o$ new samples from the candidate set based on residuals to replace part of the current data, keeping a constant training sample size.
    \item Volume Sampling for Streaming Active Learning (VeSSAL) \cite{saran2023streaming}: Selects samples based on their gradient distribution relative to the model, which is a type of diversity-based sampling in AL.
\end{itemize}

\subsection{Implementation}
% Modified [1]
Our algorithm was implemented using the PyTorch framework on a single NVIDIA A6000 GPU. 
We utilized $N=10$k samples for initial training and 5k samples for validation for each experiment. 
% For the testing set, if there are too few samples, random sampling may fail to capture representative oscillatory samples, potentially affecting our evaluation. Therefore, 
For a thorough evaluation, our testing data consists of four subsets, characterizing different types of the coefficient domain: overall (entire testing data), majority (non-oscillatory samples only), minority (oscillatory samples only), and boundary (top 20\% samples ranked by gradient using Eq. \ref{samples_gradient}).
The total size of the testing data varies between 7k--60k, depending on the oscillatory systems.  

The neural network (Multi-Layer Perceptron), consisting of 3 or 4 hidden layers, was trained for 3k epochs per sampling cycle using the Adam optimizer with a learning rate of 2--2.5$\times10^{-3}$, employing full batch training and early stopping. For key hyperparameters, the GF filtering ratio was set to $r=20\%$, with $K=5$ nearest neighbors and a GF sample size of $n_f=N/2$. During the sampling cycles, the MGS ratio was set to $n_{v1}:n_{v2}=6:4$, with an MGS sample size of $n_s=N/2$. Implementation details and other parameters for each experiment can be found in the Appendix.
All reported results below were based on five independent experiments.
\begin{table*}[!tb]
  \centering
  \caption{Imbalance Ratio (IR) and Gini Index (GI) for five baseline methods and our HGGS across four biological systems. HGGS achieves the lowest IR and GI in most cases. IS and IS$^\dag$ are unquantifiable due to significant sampling fluctuations.
  % A lower IR indicates more balanced data, while a lower GI reflects greater diversity in oscillatory patterns.
  }
    \begin{tabular}{c|cc|cc|cc|cc}
    \toprule
    \multirow{2}[0]{*}{Model} & \multicolumn{2}{c|}{Brusselator System} & \multicolumn{2}{c|}{Cell Cycle System} & \multicolumn{2}{c|}{MPF System} & \multicolumn{2}{c}{Activator Inhibitor System} \\
          & IR & GI & IR & GI & IR & GI & IR &GI \\
    \midrule
    LHS   & 1.47±0.00 & 0.76±0.00 & 4.43±0.00 & 0.85±0.00 & 5.27±0.00 & 0.89±0.00 & 11.84±0.00 & 0.94±0.00 \\
    WRS & 1.37±0.05 & 0.75±0.00 & 4.41±0.12 & 0.85±0.00 & 5.16±0.18 & 0.89±0.00 & 11.32±0.48 & 0.94±0.00 \\
    VeSSAL & 1.42±0.12 & 0.75±0.01 & 3.98±0.16 & 0.84±0.01 & 3.57±0.37 & 0.85±0.01 & 10.96±0.56 & 0.94±0.00 \\
    US-S & 1.18±0.11 & 0.72±0.01 & 3.93±1.53 & 0.83±0.05 & 3.75±0.37 & 0.85±0.01 & 11.58±2.90 & 0.94±0.02 \\
    US-P & \textbf{1.08±0.17} & \textbf{0.70±0.02} & 2.3±0.22 & 0.76±0.02 & 2.75±0.41 & 0.81±0.02 & 7.12±0.24 & 0.91±0.00 \\
    Ours & 1.18±0.12 & 0.71±0.03 & \textbf{1.10±0.02} & \textbf{0.58±0.01} & \textbf{1.11±0.07} & \textbf{0.65±0.01} & \textbf{2.88±0.13} & \textbf{0.77±0.01} \\
    \bottomrule
    \end{tabular}%
  \label{tab:SamplingMetrics}%
\end{table*}%

\begin{figure*}[!tb]
    \subfigure[Brusselator System]{
    \includegraphics[width=\linewidth]{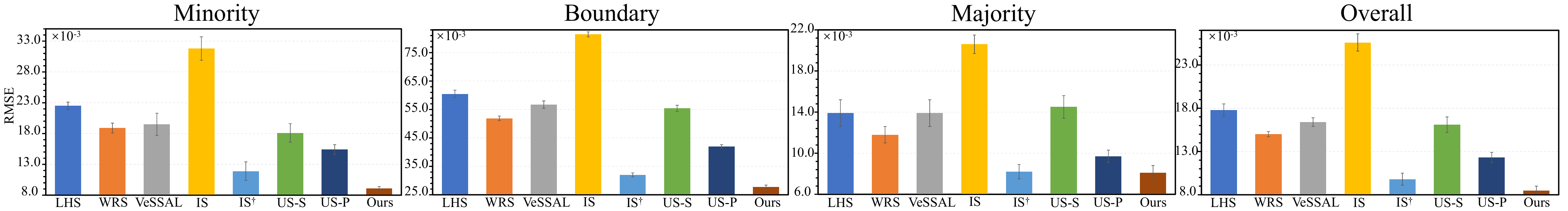}
    \label{subfig:Brusselator}
    }
    \subfigure[Cell Cycle System]{
    \includegraphics[width=\linewidth]{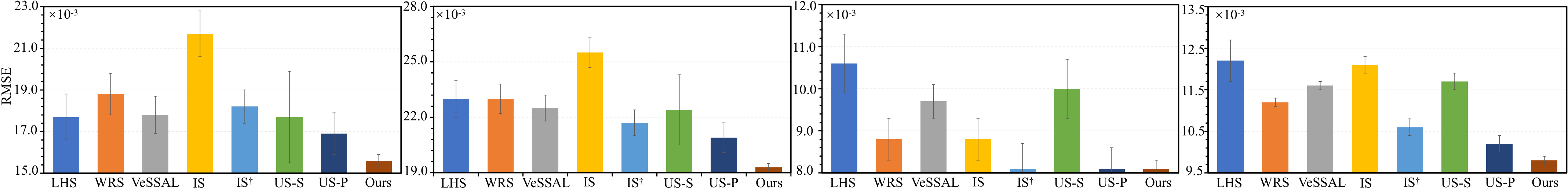}
    \label{subfig:CellCycle}
    }
    \subfigure[MPF System]{
    \includegraphics[width=\linewidth]{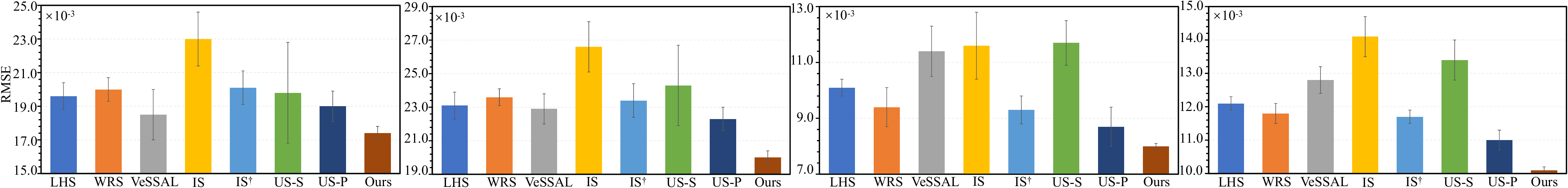}
        \label{subfig:MPF}
    }
    \subfigure[Activator Inhibitor System]{
    \includegraphics[width=\linewidth]{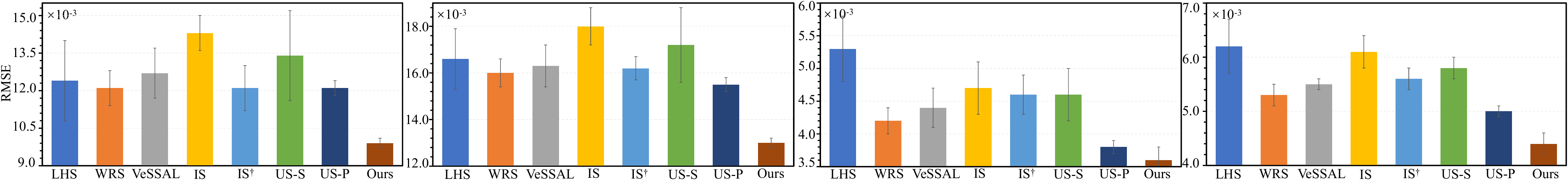}
    \label{subfig:ActivatorInhibitor}
    }
    \caption{Accuracy comparison of seven baseline methods (LHS, WRS, IS, IS$^\dag$, US-S, US-P) and our HGGS across four biological systems. HGGS obtained the lowest RMSEs across all testing subsets: minority, boundary, majority, and overall.}
    % The majority refers to the non-oscillatory class, The minority to the oscillatory class, and boundary to the top 20\% samples filtered by the gradient function in Eq. \ref{samples_gradient}.
\end{figure*}

\subsection{Results}
\textbf{Metrics.}
Root Mean Square Error (RMSE) is our primary metric. Imbalance Ratio (IR) and Gini Index (GI) quantify the proportion of non-oscillatory to oscillatory samples and the diversity of oscillatory frequency labels, respectively. Lower IR and GI indicate more effective handling of non-oscillatory bias and oscillatory boundary sensitivity by the sampling method. RMSE, IR, and GI are defined below. 
%The results show the superior efficacy of our approach, detailed in the following.

\begin{equation*}\label{equ:RMSE}
\begin{aligned}
    {\rm RMSE}=\sqrt{\frac{1}{N} \sum_{i=1}^{N}\norm{f_{nn}(\bm{\lambda}^{(i)};\bm{\Theta})-y^{(i)}}_2^2}
\end{aligned}
\end{equation*}
\begin{equation*}\label{equ:IR,GI}
\begin{aligned}
    {{\rm IR}}=\frac{\sum_{i=1}^NI(y^{(i)}=0)}{\sum_{i=1}^NI(y^{(i)}\neq0)};\ 
    {{\rm GI}}=\frac{\sum_{i=1}^N\sum_{j=1}^N\abs{y^{(i)}-y^{(j)}}}{2N\sum_{i=1}^Ny^{(i)}}
    % \nonumber
\end{aligned}
\end{equation*}

\textbf{Brusselator.}
In Fig.~\ref{subfig:Brusselator}, our method achieves the lowest RMSE for the overall testing data (0.0085). Notably, for crucial minority and boundary cases, we observe 24\%--71\% and 13\%--66\% improvement over the baseline approaches.
In a low-dimensional space like this system, with densely-packed data, IS$^\dag$ (minority: 0.0119; boundary: 0.0320) performs relatively well but lacks sufficient boundary information. 
% Similarly, US-P (minority: 0.0154; boundary: 0.0420) is limited by random sampling, resulting in fewer samples in boundary regions. 
HGGS, however, effectively targets more boundary instances, particularly in high-residual boundary regions, as illustrated in Appendix. Due to the relatively mild data imbalance in the Brusselator system, the differences in IR and GI between our method, US-S, and US-P are minimal. Nonetheless, HGGS efficiently samples at the boundaries and improves prediction accuracy compared to the others.

\textbf{Cell Cycle.} 
In this complex cell cycle system, our method distinguishes itself with the lowest IR and GI in Table~\ref{tab:SamplingMetrics}, highlighting its dual strength in balancing data distribution and enriching sample diversity. HGGS also obtains the lowest RMSE error (overall: 0.0098), as shown in Fig.~\ref{subfig:CellCycle}. Specifically for minority and boundary cases, our method shows improvements of 8\%--28$\%$ and 8\%--24$\%$ over the other seven baselines. 
US-P (minority: 0.0169; boundary: 0.0209) includes high-residual samples but suffers from redundancy due to random sampling. In contrast, our method directly targets high-risk areas, avoiding this redundancy and demonstrating superior efficiency and effectiveness.

\textbf{MPF.}
Fig.~\ref{subfig:MPF} shows that our method excels in achieving minimal RMSE (overall: 0.0101), a decrease of 7\%--27$\%$ compared to other methods.
It also stands out for minority and boundary cases (minority: 0.0174; boundary: 0.0200).
VeSSAL performs adequately for minority and boundary cases, but its weaker performance in the majority case reduces its overall effectiveness.
Table~\ref{tab:SamplingMetrics} further underscores the effectiveness of HGGS in mitigating non-oscillatory bias in the MPF system. 
Our method achieves the lowest GI among all baselines, reflecting its superior ability to uncover informative oscillatory patterns and enhance diversity. 
%This highlights the dual capabilities of our method: managing data imbalance and improving oscillation information quality, leading to a more comprehensive dataset representation.

% 1x4 version
% \begin{figure*}[!tb]
%     \includegraphics[width=\linewidth]{Paper/figures/AblationStudy.pdf}
%     \caption{Ablation study for majority and minority classes across four oscillatory systems. Both Gradient-based Filtering (GF) and Genetic Sampling (GS) layers contribute to the improvement in model accuracy.}
%     \label{fig:AblationStudy}
% \end{figure*}

\textbf{Activator Inhibitor.}
% Ours: Overall=0.0044
% improvements: overall=12%~29%, minority=18%~31%, boundary=16%~28%
The Activator Inhibitor system is characterized by a pronounced data imbalance, with a staggering IR of 11.84 under LHS. However, as highlighted in Table~\ref{tab:SamplingMetrics}, our method exhibits remarkable resilience against such extreme scenarios, effectively reducing the IR to about 2.88 while simultaneously enhancing data diversity to a GI of 0.77. 
Moreover, HGGS has the lowest error across all categories (minority: 0.0099; boundary: 0.0130; majority: 0.0036; overall: 0.0044) (Fig.~\ref{subfig:ActivatorInhibitor}). It particularly excels in minority and boundary cases, showing improvements of 18\%--31$\%$ and 16\%--28$\%$ over other baselines.

% 2x2 version
\begin{figure}[!tb]
    \includegraphics[width=\linewidth]{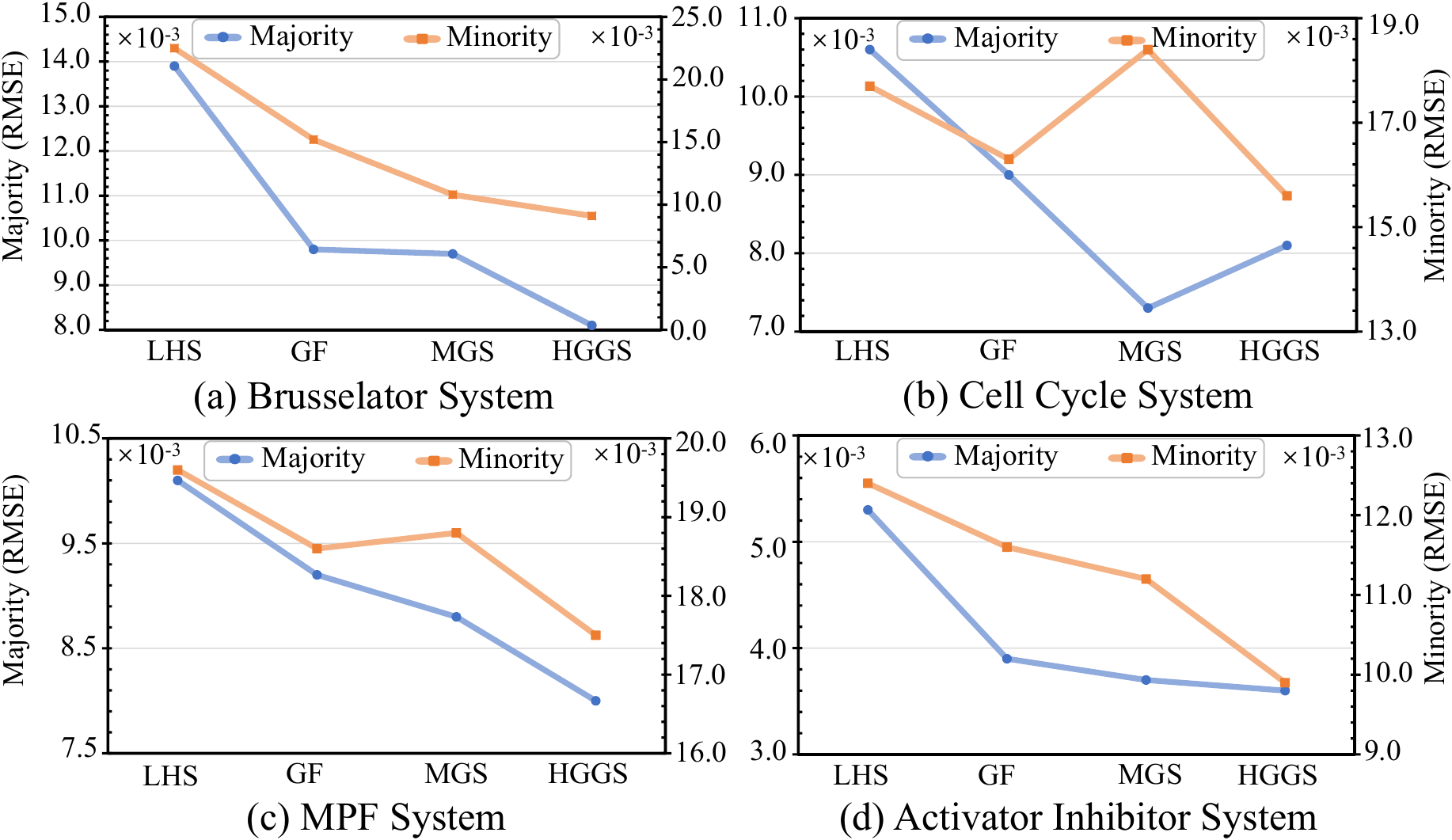}
    \caption{Ablation study for majority and minority classes across four oscillatory systems. Both Gradient-based Filtering (GF) and Multigrid Genetic Sampling (MGS) layers contribute to the improvement in model accuracy.}
    \label{fig:AblationStudy}
\end{figure}
\begin{figure}[!tb]
    \includegraphics[width=\linewidth]{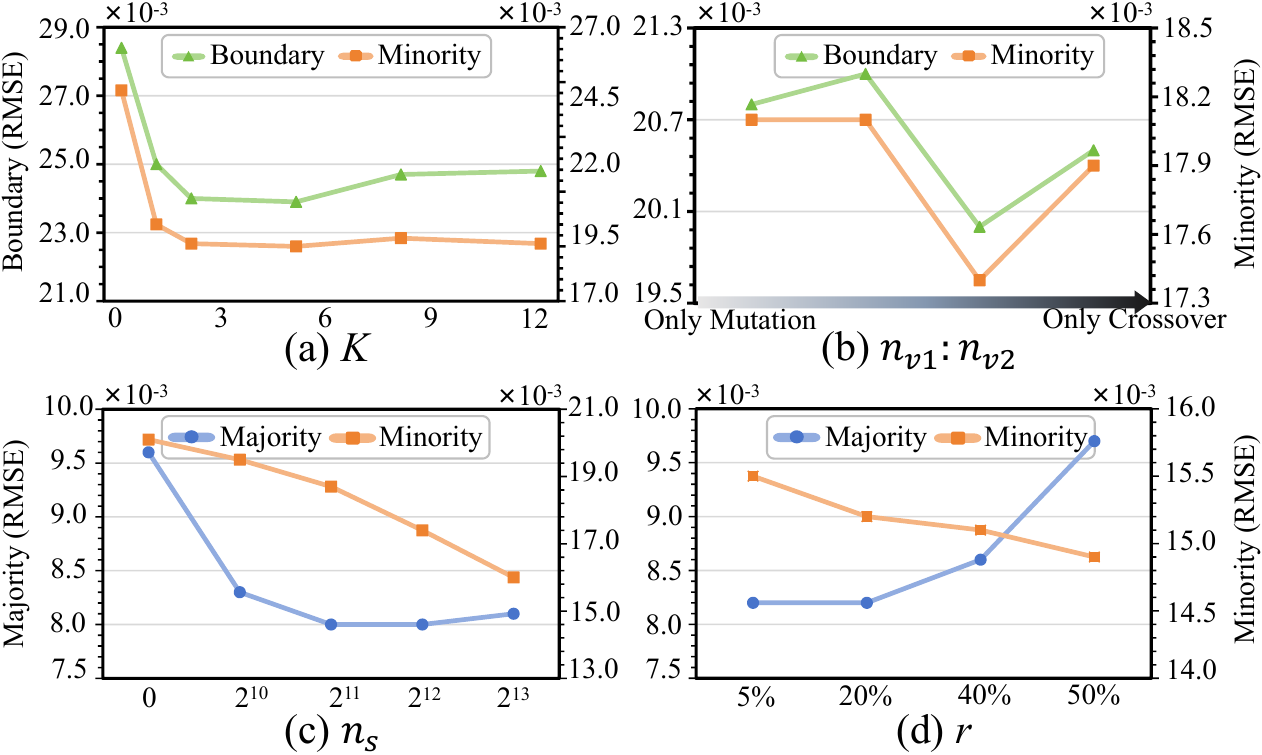}
    \caption{Sensitivity analysis on (a) number of neighbors $K$, (b) MGS ratio $n_{v1}:n_{v2}$, (c) MGS sampling size $n_s$, and (d) GF filtering ratio $r$. (a-c) are performed on the MPF system, while (d) is conducted on the Cell Cycle system.}
    \label{fig:ParameterSensitivity}
\end{figure}
\subsection{Ablation Study}

We conducted an ablation study on four biological systems to evaluate the efficacy of the two layers: Gradient-based Filtering and Multigrid Genetic Sampling. 

\textbf{Effect of Gradient-based Filtering}. Relying solely on GF followed by random sampling does not effectively address non-oscillatory bias and oscillatory boundary sensitivity. Although GF offers initial benefits, subsequent random sampling fails to sustain improvements. As shown in Fig.~\ref{fig:AblationStudy}, GF yields higher RMSEs for majority and minority cases across four systems than HGGS, indicating that MGS produces more informative samples than random sampling.

\textbf{Effect of Multigrid Genetic Sampling}. When MGS is used without prior GF, it struggles to precisely identify informative minority samples, leading to higher RMSEs for minority class across all four systems compared to HGGS (Fig.~\ref{fig:AblationStudy}). This suggests that GF provides critical guidance for MGS to effectively target and refine minority samples.

These findings underscore the synergy between GF and MGS within HGGS. GF enables an efficient extraction of sample domains, identifying coarse boundaries rich in critical information. Informed by GF, MGS operates with greater precision to refine minority samples and explore new high-risk areas. Together, they enhance data representativeness and diversity, improving model performance.

\subsection{Sensitivity Analysis}

% **Table X** elucidates the impact of various Gradient-based Filtering hyperparameters, including the number of filtering samples and the count of nearest neighbors $K$ utilized for sample gradient computation, alongside Genetic-Sampling hyperparameters, on the performance of the HGGS methodology.
% The empirical findings reveal that when the Gradient-based Filtering component yields predominantly boundary information due to an excessively large $S_{\Omega_{gd1}}$ setting, the model becomes overly fixated on minority samples. Consequently, this leads to a degradation in performance concerning the Non-Oscillation category, illustrating the importance of a balanced approach in capturing both majority and minority class characteristics. In this condition, preserving a portion of Non-Oscillation samples that are beneficial to the model ensures a more comprehensive understanding of the data landscape.

Fig.~\ref{fig:ParameterSensitivity}(a-c) illustrates the effects of key model parameters, including the number of neighbors $K$, MGS ratio $n_{v1}:n_{v2}$, and MGS sampling size $n_s$ on the MPF system. In Fig.~\ref{fig:ParameterSensitivity}(a), varying $K$ produces an elbow curve, with $K=5$ offering optimal performance. Increasing $K$ beyond this point has little effect on gradient estimation, while decreasing $K$ leads to significant errors in oscillatory frequency estimation.
In Fig.~\ref{fig:ParameterSensitivity}(b), MGS cannot achieve optimal performance using only crossover or mutation operation. Our experiments found that an MGS ratio of 6:4 yielded the lowest RMSE. 
In Fig.~\ref{fig:ParameterSensitivity}(c), increasing the MGS sampling size $n_s$ continues to reduce RMSE for the minority case, while the majority case shows no further improvement beyond 2k new samples. 
Fig.~\ref{fig:ParameterSensitivity}(d) shows the effect of GF filtering ratio on the Cell Cycle system. Increasing $r$ reduces the sampling of majority (non-oscillation) class, leading to higher error, while the opposite is true for minority class. A ratio of $r=20\%$ offers the best balance between majority and minority samples.
\section{Conclusion}\label{sec:Conclusion}
This paper introduces Hierarchical Gradient-based Genetic Sampling, a two-layer framework designed to address non-oscillatory bias and boundary sensitivity in predicting biological oscillations. The first layer, gradient-based filtering, selects a representative subset from initial random sampling, creating a balanced coarse dataset. The second layer, genetic sampling, refines minority instances and explores new high-residual information, enhancing data diversity. Experiments show that HGGS achieves the best accuracy across four biological systems, particularly for the oscillation class.

\newpage
\bibliography{aaai25}
% \newpage
% \clearpage
\iffalse
\input{ReproducibilityChecklist}
\fi
% \iffalse
\newpage
\appendix
\setcounter{figure}{0}
\renewcommand{\thefigure}{A\arabic{figure}}
\setcounter{table}{0}
\renewcommand{\thetable}{A\arabic{table}}

\section*{A Appendix}
\subsection{A.1 Biological System Dataset}
This section provides a detailed description of the biological systems studied, along with the corresponding ordinary differential equations (ODEs) that govern their dynamics. Table~\ref{tab:expsettings} lists the simulation settings for generating benchmark data of each system, including time span, initial conditions, and ranges of system coefficients.

% Table generated by Excel2LaTeX from sheet 'Baseline'
\begin{table*}[!tb]
  \centering
  \caption{Simualtion settings for four biological systems.}
    \begin{tabular}{c|c|c|c|c}
    \toprule
    Attributes/Biological System & Brusselator & Cell Cycle & MPF   & Activator Inhibitor \\
    \midrule
    Time span & [0,500] & [0,1000] & [0,1000] & [0,5000] \\
    \midrule
    \multirow{5}{*}{Initial condition} 
    & 
    $\langle X \rangle|_{t=0}=10$, & 
    $V|_{t=0}=30$, & 
    $\langle X \rangle|_{t=0}=0.03657$, & 
    $\langle X \rangle|_{t=0}=1$, \\
    & &
    $\langle X \rangle|_{t=0}=320$, & 
    & \\
    & &
    $\langle Y_T \rangle|_{t=0}=100$, & 
    & \\
    & &
    $\langle Y \rangle|_{t=0}=100$, &  
    & \\
    & 
    $\langle Y \rangle|_{t=0}=10$. & 
    $\langle Z \rangle|_{t=0}=200$. & 
    $\langle Y \rangle|_{t=0}=0.36615$. & 
    $\langle Y \rangle|_{t=0}=4$. \\
    \midrule
    \multirow{6}{*}{System coefficient range} 
    &
    $\lambda_1\in[0,5.0]$, & 
    $\lambda_1\in[0,15.3]$, &
    $\lambda_1\in[0,0.1]$, &
    $\lambda_1\in[0,28.0]$,\\
    & &
    $\lambda_2\in[0,0.4]$, &
    $\lambda_2\in[0,0.1]$, &
    $\lambda_2\in[0,1.0]$,\\
    & & 
    $\lambda_3\in[0,13.5]$, &
    $\lambda_3\in[0,0.4]$, &
    $\lambda_3\in[0,1.0]$,\\
    & & 
    $\lambda_4\in[0,0.2]$, &
    $\lambda_4\in[0,15.0]$, &
    $\lambda_4\in[0,10.0]$,\\
    & & 
    $\lambda_5\in[0,13.5]$, &
    $\lambda_5\in[0,1.0]$, &
    $\lambda_5\in[0,50.0]$,\\
    &
    $\lambda_2\in[0,15.0]$. & 
    $\lambda_6\in[0,1.0]$. &
    $\lambda_6\in[0,10.0]$. &
    $\lambda_6\in[0,10.0]$.\\
    % \midrule
    % MLP Architecture & [128, 256, 128] & [128, 128, 128, 128] & [256, 256, 256, 256] \\
    \bottomrule
    \end{tabular}%
  \label{tab:expsettings}%
\end{table*}%

% \begin{figure*}[!tb]
%     \includegraphics[width=\linewidth]{Paper/figures/[Appendix]HGGS-Sampling.pdf}
%     \caption{Oscillatory frequency distribution of samples obtained by HGGS. (a) shows the initial dataset from Latin Hypercube Sampling (LHS). (b) presents the coarse data extracted from (a) using Gradient-based Filtering (GF) technique. Subsequently, (c-d) show refined sampling at boundary regions based on the information from (b) using Multigrid Genetic Sampling strategy. The black dashed box illustrates the capability of HGGS to sample extensively at boundaries.}
%     \label{fig-Appendix:Brusselator-HGGS-Sampling}
% \end{figure*}

% Brusselator
% The Brusselator system is a two-variable system used to describe oscillatory chemical reactions. 
\textbf{Brusselator System \cite{prigogine1978time}:} 
The Brusselator system is renowned for its ability to generate periodic oscillations under specific coefficient conditions, making it a valuable model for understanding chemical oscillations and wave patterns. Its versatility has led to applications in diverse fields, including economics, biology, and management systems. The model describes a chemical reaction system involving two main species, typically denoted as $\langle X \rangle$ and $\langle Y \rangle$. These species interact through a set of reactions that can be simplified and represented by the following ODEs:
$$
\left\{
\begin{aligned}
    \dv{}{t}\langle X \rangle &= \lambda_1 - (\lambda_2+1)\langle X \rangle + \langle X \rangle^2\times\langle Y \rangle \\
    \dv{}{t}\langle Y \rangle &= \lambda_2\langle X \rangle - \langle X \rangle^2\times\langle Y \rangle \\
\end{aligned}
\right.
$$
The system can exhibit either a stable equilibrium or a limit cycle (sustained oscillations), depending on the values of system coefficients ($\lambda_1$ and $\lambda_2$). By varying these coefficients, various oscillatory behaviors can be simulated, representing different types of chemical reaction dynamics. 
%The system's periodic behavior is analyzed over a time period of $t\in[0,500]$.

% Cell Cycle System
\textbf{Cell Cycle System \cite{liu2012hybrid}:} 
The cell cycle system is essential for eukaryotic cell proliferation. The progression through the cell cycle's phases--G1, S, G2, and M--is governed by a complex interplay of regulatory proteins, including cyclins and cyclin-dependent kinases (CDKs). Cdk1, in combination with cyclins like Clb2, drives the cell through mitosis, while Cdc20 and Cdc14 regulate transitions between cell cycle phases. The cell cycle follows a rhythmic pattern, characterized by the periodic activation and inactivation of these regulatory proteins. This oscillatory behavior ensures precise transitions between phases, maintaining accurate timing for cell growth, DNA replication, and division.
The dynamics of the cyclins and CDKs are governed by

$$
\left\{
\begin{aligned}
  \dv{}{t}V &= 0.006  V,\\ 
  \dv{}{t}\langle X \rangle &= \lambda_1 \left(\frac{1.04 V}{3.5}\right) V - \lambda_2 \langle X \rangle \\&- 0.00741 \frac{\langle X \rangle \langle Y \rangle}{V}, \\
  \dv{}{t}\langle Y_T \rangle &= \lambda_3 \left(\frac{7.0}{3.5}\right) V - \lambda_4 \langle Y_T \rangle, \\
  \dv{}{t}\langle Y \rangle &= \lambda_3 \left(\frac{7.0}{3.5}\right) V - \lambda_4 \langle Y \rangle \\&+ \frac{(29.7 V + 7.5 \langle Z \rangle) (\langle Y_T \rangle - \langle Y \rangle)}{(5.4 V + \langle Y_T \rangle - \langle Y \rangle)} \\&- \frac{1.88 \langle X \rangle \langle Y \rangle}{(5.4 V + \langle Y \rangle)}, \\
  \dv{}{t}\langle Z \rangle &= \lambda_5 \left(\frac{0.001 + 10 \frac{\langle X \rangle^2}{(756 V)^2 + \langle X \rangle^2}}{0.15}\right) V \\&- \lambda_6 \langle Z \rangle
\end{aligned}
\right.
$$
The average number of Cdk1 is denoted as $\langle X \rangle$, Clb2 by $\langle Y \rangle$, and $\langle Z \rangle$ represents the composite species (Cdc20 and Cdc14). The term $\langle Y_T \rangle$ refers to the total average number of phosphorylated and unphosphorylated Clb2, while $V$ stands for cell volume. The reaction rates are governed by the coefficients $\lambda_1$ and $\lambda_2$ for $X$, $\lambda_3$ and $\lambda_4$ for $Y$, and $\lambda_5$ and $\lambda_6$ for $Z$.
% \chen{I don't understand.} The dynamic system involves a total of 5 variables, our analysis zeroes in on the periodic behavior exhibited by 3 particular variables, denoted as $\langle X \rangle,\langle Y \rangle$, and $\langle Z \rangle$. 

%The initial conditions for this system are set as $\bm{u}(0,\bm{\lambda})=[30, 320, 100, 100, 200]$, and the examination of its periodic nature is carried out over the extended period from $t=0$ to $t=1000$.

% MPF System
\textbf{MPF \cite{novak1993modeling}:} 
% Mitotic Promoting Factor (MPF) plays a pivotal role in the cell division cycle. MPF is crucial for the transition from the G2 phase to mitosis, ensuring proper chromosome segregation and cytokinesis. \chen{Explain the oscillation pattern.} The dynamic system of MPF is characterized by two variables, \chen{add description for x, y.} $\langle X \rangle$ and $\langle Y \rangle$, described below.
Mitotic Promoting Factor (MPF) is crucial for cell division, regulating the transition from the G2 phase to the M phase to ensure accurate chromosome segregation and cytokinesis. MPF, composed of cyclin and Cdc2, is activated by phosphorylation to drive cell division and then disassembles, with cyclin degraded and Cdc2 recycled for the next cycle, creating a periodic pattern.
The dynamics of MPF are described by two variables, $\langle X \rangle = \frac{[Active\ MPF]}{[Total\ cdc2]}$ and $\langle Y \rangle = \frac{[Total\ cyclin]}{[Total\ cdc2]}$, where brackets $[~]$ denote the concentration of corresponding species. 
$$
\left\{
\begin{aligned}
  \dv{}{t} \langle X \rangle &= \frac{\lambda_1}{G} - (\lambda_2 + 10 \langle X \rangle^2 + \lambda_4) \times \langle X \rangle \\&+    (\lambda_3 + 100 \langle X \rangle^2) \left( \frac{\langle Y \rangle}{G} - \langle X \rangle \right) \\
  \dv{}{t} \langle Y \rangle &= \lambda_1 - (\lambda_2 + 10 \langle X \rangle^2) \times \langle Y \rangle\\
   G&=1+\frac {\lambda_5} {\lambda_6}.
\end{aligned}
\right.
$$
Here, $\lambda_1$, $\lambda_2$, and $\lambda_3$ represent the effects of active MPF on cyclin degradation and tyrosine dephosphorylation, while $\lambda_4$, $\lambda_5$, and $\lambda_6$ denote the enzymatic action rates of Wee1, INH, and CAK, respectively. The ratio $G$ (composed of $\lambda_5$ and $\lambda_6$) influences the phosphorylation state of Cdc2, specifically affecting tyrosine-15 and threonine-167.
% \chen{Explain G, coefficients, add oscillatory frequency}

% Activator Inhibitor System
\textbf{Activator Inhibitor System \cite{murray2002mathematical}:} This model describes how two interacting substances--an activator and an inhibitor--regulate processes such as enzyme activity, chemical reactions, and pattern formation. It captures various dynamic behaviors, including oscillations and spatial patterns. This system studied here includes two variables: the activator $\langle X \rangle$ and the inhibitor $\langle Y \rangle$. The rate at which these concentrations change over time is influenced by the system coefficients $\lambda_1$ through $\lambda_6$.  
The governing ODEs are shown below.
$$
\left\{
\begin{aligned}
  \dv{}{t} \langle X \rangle&= \frac{\lambda_4 + \lambda_5 \langle X \rangle^2}{1 + \langle X \rangle^2 + \lambda_6 \langle Y \rangle} - \langle X \rangle \\
  \dv{}{t} \langle Y \rangle&= \lambda_3 (\lambda_1 \langle X \rangle + \lambda_2 - \langle Y \rangle),
\end{aligned}
\right.
$$
% where the Activator Inhibitor system, also defined by 2 variables $\langle X \rangle$ and $\langle Y \rangle$, is initialized under the condition $\bm{u}(0,\bm{\lambda})=[1, 4]$. Our analysis of its periodic behavior encompasses a broader time frame, extending from $t=0$ to $t=5000$. 

\subsection{A.2 Implementation Details}
The code and data are available in the supplementary materials. Our algorithm was implemented using the PyTorch framework on a single NVIDIA A6000 GPU. Alongside $N=10$k samples for initial training and 5k samples for validation for each experiment, a comprehensive testing dataset for thoroughly evaluating performance was prepared.
The test dataset sizes for the Brusselator, Cell Cycle, MPF, and Activator-Inhibitor systems are 7.5k, 45k, 51k, and 51k, respectively. These are further divided into majority subsets with 4.5k, 37k, 43k, and 47k samples; minority subsets with 3k, 8k, 8k, and 4k samples; and boundary subsets with 0.6k, 5k, 7k, and 3k samples, respectively.

To balance the number of samples generated by different methods and enable a fair comparison, we introduced the definition of sampling efficiency $\eta$, defined as follows:
\begin{equation*}
\begin{aligned}
    \eta=\frac {|S_{\Omega}^{(m_c)}|} {|S_{\Omega}| + |S_{\Omega_{gs}}|}
\end{aligned}
\end{equation*}
where $m_c$ denotes the number of sampling iterations, $S_{\Omega}^{(m_c)}$ represents the total training samples, $S_\Omega$ refers to the initial dataset, and $S_{\Omega_{gs}}$ represents the new samples obtained through the sampling method. For the LHS, IS, and IS$^\dag$ methods, $S_{\Omega_{gs}} = 0$ and $S_{\Omega}=10000$, while the sampling efficiency of other methods are set at $\frac{2}{3}$.

For model training, we use a Multi-Layer Perceptron (MLP) with the Adam optimizer and an exponential learning rate scheduler, along with full-batch training and early stopping. Other parameters used for model training are listed in Table~\ref{tab:Implemention Detail}. For key hyperparameters, the GF filtering ratio was set to $r=20\%$, with $K=5$ nearest neighbors and a GF sample size of $n_f=N/2$. During the sampling cycles, the MGS ratio was set to $n_{v1}:n_{v2}=6:4$, with an MGS sample size of $n_s=N/2$.

\begin{figure*}[!tb]
    \includegraphics[width=\linewidth]{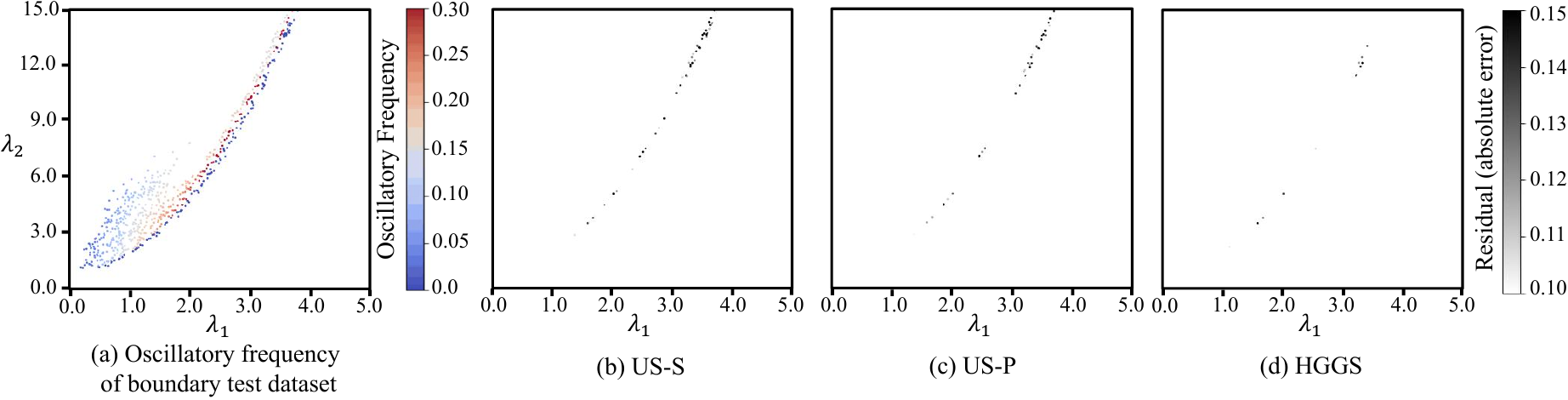}
    \caption{Performance comparison (residual distributions) between US-S, US-P, and our HGGS methods on the boundary test dataset of the Brusselator system.}
    \label{fig-Appendix:Brusselator-HGGS-Compare-MAE}
\end{figure*}
% Results of the boundary test dataset experiments on the Brusselator system. (a) shows the oscillatory frequency distribution of the boundary test dataset, while (b-c) illustrate the residual distribution for the boundary test dataset using US-S, US-P, and HGGS.
\subsection{A.3 Boundary Analysis on Brusselator System}
As a complementary to the main results, we demonstrate the effectiveness of our method using the Brusselator system, which features only two coefficients and allows for clear visualization. Fig.~\ref{fig-Appendix:Brusselator-HGGS-Compare-MAE}(a) depicts the boundary distribution of oscillatory frequency within the Brusselator system. In Fig.~\ref{fig-Appendix:Brusselator-HGGS-Compare-MAE}(b-c) our HGGS outperforms US-P and US-S, achieving significantly lower absolute errors in the boundary region, as indicated by fewer black dots.
% In this section, we demonstrate the effectiveness of our method on a Brusselator system with only two coefficients. Fig.~\ref{fig-Appendix:Brusselator-HGGS-Sampling}(a-d) illustrates the sampling distribution of HGGS from the Gradient-based Filtering (GF) phase to the Multigrid Genetic Sampling (MGS) phase. Fig.~\ref{fig-Appendix:Brusselator-HGGS-Sampling}(b) demonstrates that our method can filter out informative data from Fig.~\ref{fig-Appendix:Brusselator-HGGS-Sampling}(a) during the GF phase, particularly by removing unnecessary non-oscillatory information. Subsequently, as the number of MGS samples increases, HGGS continues to sample more from the boundary (indicated by the black dashed box in Fig.~\ref{fig-Appendix:Brusselator-HGGS-Sampling}(c-d)). In Fig.~\ref{fig-Appendix:Brusselator-HGGS-Compare-MAE}, we compare our method HGGS with US-P and US-S, highlighting that HGGS achieves a lower RMSE in the boundary region (black dashed box in Fig.~\ref{fig-Appendix:Brusselator-HGGS-Compare-MAE}(b)) by sampling more effectively. In contrast, both US-S and US-P rely on random sampling for new samples, making it difficult for them to accurately capture boundary samples. This results in higher errors in the boundary region due to the inclusion of many redundant non-oscillatory samples.

\begin{table*}[htbp]
  \centering
  \caption{Experimental details for four biological systems. The epoch refers to the number of training epochs for coarse data generated by the GF layer, and $m_e$ represents the number of training epochs for new data following each sampling by the MGS layer.}
    \begin{tabular}{c|c|c|c|c}
    \toprule
    Biological System & \multicolumn{1}{c|}{Brusselator} & \multicolumn{1}{c|}{Cell Cycle} & \multicolumn{1}{c|}{MPF} & \multicolumn{1}{c}{Activator Inhibitor} \\
    \midrule
    N     & \multicolumn{1}{c|}{10000} & \multicolumn{1}{c|}{10000} & \multicolumn{1}{c|}{10000} & \multicolumn{1}{c}{10000} \\
    % \midrule
    learning rate    & \multicolumn{1}{c|}{$2\times10^{-3}$} & \multicolumn{1}{c|}{$2.5\times10^{-3}$} & \multicolumn{1}{c|}{$2\times10^{-3}$} & \multicolumn{1}{c}{$2\times10^{-3}$} \\
    % \midrule
    weight decay & \multicolumn{1}{c|}{$1\times 10^{-5}$} & \multicolumn{1}{c|}{$1\times 10^{-5}$} & \multicolumn{1}{c|}{$1\times 10^{-5}$} & \multicolumn{1}{c}{$1\times 10^{-5}$} \\
    % \midrule
    warm epoch/epoch & \multicolumn{1}{c|}{300/3000} & \multicolumn{1}{c|}{200/2000} & \multicolumn{1}{c|}{300/3000} & \multicolumn{1}{c}{250/2500} \\
    % \midrule
    $m_c$/$m_e$ & \multicolumn{1}{c|}{20/3000} & \multicolumn{1}{c|}{20/3000} & \multicolumn{1}{c|}{20/3000} & \multicolumn{1}{c}{20/3000} \\
    % \bottomrule
    MLP architecture & [128,256,128] & [128,256,128] & [128,128,128,128] & [256,256,256,256] \\
    \bottomrule
    \end{tabular}%
  \label{tab:Implemention Detail}%
\end{table*}%

\subsection{A.4 Proof of Statistical Error Reduction via Residual Distribution Sampling}
Here, we present a proof demonstrating how sampling based on the residual distribution can reduce statistical error, as stated in the Preliminary. Our goal is to minimize the statistical error, represented by the first term in the following inequality:
\begin{equation*}
\begin{aligned}
    E(||f_{nn}(\cdot)_N^* - y(\cdot)||_\Omega)
    \leq &E(||f_{nn}(\cdot)_N^* - f_{nn}(\cdot)^*||_\Omega)\\
    &+ ||f_{nn}(\cdot)^* - y(\cdot)||_\Omega,
\end{aligned}
\end{equation*}
where $f_{nn}(\cdot)_N^*$ is the model obtained from the Monte Carlo approximation of the loss function $L_N$. The statistical error is directly related to the variance of the stochastic gradient,  $\mathrm{Var}[\nabla_{\Theta} L_N]$. Therefore, reducing this variance is crucial for accurate gradient estimation and, consequently, for minimizing statistical error. The loss functions $L$ and $L_N$ are defined as follows:
\begin{equation*}
\begin{aligned}
L(f_{nn}(\bm{\lambda}; \bm{\Theta}), y)&= \norm{l(f_{nn}(\bm{\lambda}; \bm{\Theta}), y)}_{2, \Omega}^2,\\
L(\bm{\Theta}) \approx L_N(\bm{\Theta})
&= \norm{l(f_{nn}(\bm{\lambda}; \bm{\Theta}), y)}_{2, S_\Omega}^2 \\
&= \frac{1}{N} \sum_{i=1}^{N} \norm{l(f_{nn}(\bm{\lambda}^{(i)}; \bm{\Theta}), y^{(i)})}_2^2,
\end{aligned}
\end{equation*}
where $N$ is the number of samples and $S_\Omega = \{\bm{\lambda}^{(i)}\}_{i=1}^{N}$ represents the set of sampled coefficients from the domain $\Omega$. The residual $l(f_{nn}(\bm{\lambda}; \bm{\Theta}), y) = \abs{f_{nn}(\bm{\lambda}; \bm{\Theta}) - y}$, and $\norm{\cdot}_2$ denotes Euclidean norm.

Since our method samples according to the distribution $p$, the loss function can be rewritten as:
\begin{equation*}\label{loss function(continue)-ours}
L(f_{nn}(\bm{\lambda}; \bm{\Theta}), y) = \norm{p \cdot l(f_{nn}(\bm{\lambda}; \bm{\Theta}), y)}_{2, \Omega}^2,
\end{equation*}
where $p$ is our sampling distribution. Sampling according to distribution $p$ yields an unbiased stochastic gradient $\frac{1}{Np} \nabla_{\Theta} L_N$. Therefore, based on \cite{lu2023pa}, to minimize the variance $\mathrm{Var}_p[\frac{1}{Np}\nabla_{\Theta} L_N]$, the optimization objective becomes:
\begin{equation*}
\min_p E_p \left[ \frac{1}{(Np)^2} \norm{\nabla_{\Theta} L_N}^2 \right] = \sum_{i=1}^N \frac{1}{N^2 p^{(i)}} \norm{\nabla_{\Theta} L_N^{(i)}}^2
\end{equation*}
subject to $\sum_{i=1}^N p^{(i)} = 1$ and $p^{(i)} \geq 0$ for all $i \in [1, N]$.

Using Lagrange multipliers, the optimal distribution $p$ is given by:
\begin{equation*}
p^{(i)} = \frac{\nabla_{\Theta} L_N^{(i)}}{\sum_{i=1}^N \nabla_{\Theta} L_N^{(i)}}
\end{equation*}
Since $\sup\{\nabla_{\Theta} L_N\} \leq \nabla_{L}$, where $\nabla_{L}$ represents the derivative of $L_N$ with respect to the output of the last layer of the model and $\nabla_{L}\sim l$, we obtain
\begin{equation*}
p^{(i)} = \frac{l^{(i)}}{\sum_{i=1}^N l^{(i)}},
\end{equation*}
demonstrating that sampling based on the residual distribution can reduce statistical error.
% \fi

\end{document}